\documentclass[letterpaper, 10 pt, conference]{ieeeconf}  

\IEEEoverridecommandlockouts                              

\usepackage[ruled, vlined, linesnumbered]{algorithm2e}
\usepackage[utf8]{inputenc}
\usepackage[english]{babel}
\usepackage{mathptmx} 
\usepackage{amsmath} 
\usepackage{amssymb}
\usepackage{amsfonts}
\usepackage{mathtools}

\newtheorem{proposition}{Proposition}
\usepackage{bm}
\usepackage{cite}
\usepackage{comment}
\allowdisplaybreaks

\newcommand{\xu}[1]{{\color{black}{#1}}}

\usepackage{gensymb}

\newtheorem{problem}{Problem}
\usepackage[pdftex]{graphicx} 
\usepackage{caption}
\usepackage{subcaption}
\usepackage{epstopdf}
\usepackage{hyperref} 
\usepackage[capitalise]{cleveref}
\usepackage{color}
\usepackage{xcolor}
\definecolor{Questions}{HTML}{1F77B4}

\DeclareFontFamily{OMX}{yhex}{}
\DeclareFontShape{OMX}{yhex}{m}{n}{<->yhcmex10}{}
\DeclareSymbolFont{yhlargesymbols}{OMX}{yhex}{m}{n}
\DeclareMathAccent{\wideparen}{\mathord}{yhlargesymbols}{"F3}

\usepackage{algorithm2e}
\SetKwComment{Comment}{/* }{ */}

\usepackage{cancel}

\usepackage{amssymb}

\usepackage{tikz}
\usetikzlibrary{calc}
\usetikzlibrary{arrows,positioning} 
\usetikzlibrary{intersections, patterns}
\usetikzlibrary{shapes}
\usepackage{mathtools}
\usepackage{graphicx}
\usepackage{bbm}
\usepackage[cmtip,all]{xy}
\newcommand{\longsquiggly}{\xymatrix{{}\ar@{~>}[r]&{}}}

\DeclareMathAlphabet{\mathcal}{OMS}{cmsy}{m}{n}

\title{\LARGE \bf Robust Localization of Aerial Vehicles via \\ Active Control of Identical Ground Vehicles}

\author{Igor Spasojevic*, Xu Liu*, Ankit Prabhu, Alejandro Ribeiro, George J. Pappas, Vijay Kumar
\thanks{*Equal contribution. 
This work was supported by The Institute for Learning-Enabled Optimization at Scale (TILOS) funded by the National Science Foundation (NSF) under NSF Grant CCR-2112665, IoT4Ag ERC funded through NSF Grant EEC-1941529, the ARL DCIST CRA W911NF-17-2-0181, and ONR Grant N00014-20-1-2822.
All authors are with the GRASP Laboratory, {\tt\small\{igorspas, liuxu, praankit, aribeiro, pappasg, kumar\}@seas.upenn.edu}.}%
}

\newcommand\copyrighttext{%
  \footnotesize \textcopyright 2023 IEEE. Personal use of this material is permitted.
  Permission from IEEE must be obtained for all other uses, in any current or future
  media, including reprinting/republishing this material for advertising or promotional
  purposes, creating new collective works, for resale or redistribution to servers or
  lists, or reuse of any copyrighted component of this work in other works.}
\newcommand\copyrightnotice{%
\begin{tikzpicture}[remember picture,overlay]
\node[anchor=south,yshift=10pt] at (current page.south) {\fbox{\parbox{\dimexpr\textwidth-\fboxsep-\fboxrule\relax}{\copyrighttext}}};
\end{tikzpicture}%
}

\begin{document}

\maketitle
\copyrightnotice

\begin{abstract}

This paper addresses the problem of active collaborative localization in heterogeneous robot teams with unknown data association. 
It involves positioning a small number of identical unmanned ground vehicles (UGVs) at desired positions so that an unmanned aerial vehicle (UAV) can, through unlabelled measurements of UGVs, uniquely determine its global pose. 
We model the problem as a sequential two player game, in which the first player positions the UGVs and the second identifies the two distinct hypothetical poses of the UAV at which the sets of measurements to the UGVs differ by as little as possible. 
We solve the underlying problem from the vantage point of the first player for a subclass of measurement models using a mixture of local optimization and exhaustive search procedures.  
Real-world experiments with a team of UAV and UGVs show that our method can achieve centimeter-level global localization accuracy. 
We also show that our method consistently outperforms random positioning of UGVs by a large margin, with as much as a 90\% reduction in position and angular estimation error. 
Our method can tolerate a significant amount of random as well as non-stochastic measurement noise. 
This indicates its potential for reliable state estimation on board size, weight, and power (SWaP) constrained UAVs.
This work enables robust localization in perceptually-challenged GPS-denied environments, thus paving the road for large-scale multi-robot navigation and mapping.    
\end{abstract}
\section{Introduction}
Robust localization and place recognition with unknown data association is a challenging task. 
Relying purely on geometric features or landmarks in the environment can lead to erroneous results due to perceptual aliasing, changes in lighting conditions, and non-static features.
Semantic landmarks have the potential to alleviate ambiguity in data association because they are more informative and robust to viewpoint changes \cite{bowman2017probabilistic}. 
However, methods relying on natural semantic landmarks will fail if (1) there are not enough semantic objects in the field of view, (2) they form ambiguous configurations, or (3) they become unreliable for localization purposes (e.g. vehicles and humans may move, rocks on beaches may disappear with the tide, buildings may be destroyed by earthquakes). 
In this work, we consider actively positioning a small number of UGVs, which can be viewed as human-controlled semantic landmarks, into the environment to form constellations UAVs can use for robust localization.
Our previous work \cite{spasojevic2023active} addressed a version of this problem which assumed known data association and that all robots can estimate their poses in the same reference frame.
We now relax the former assumption and consider the active collaborative localization problem without prior knowledge of the UAVs' reference frame and with unknown data association.
We propose a maximin approach, in which the max step selects positions of the UGVs, and the min step estimates the data association and the pose of the UAV.

\begin{figure}[t!]
\centering
\includegraphics[trim=100 40 0 180, clip,width=0.9\columnwidth]{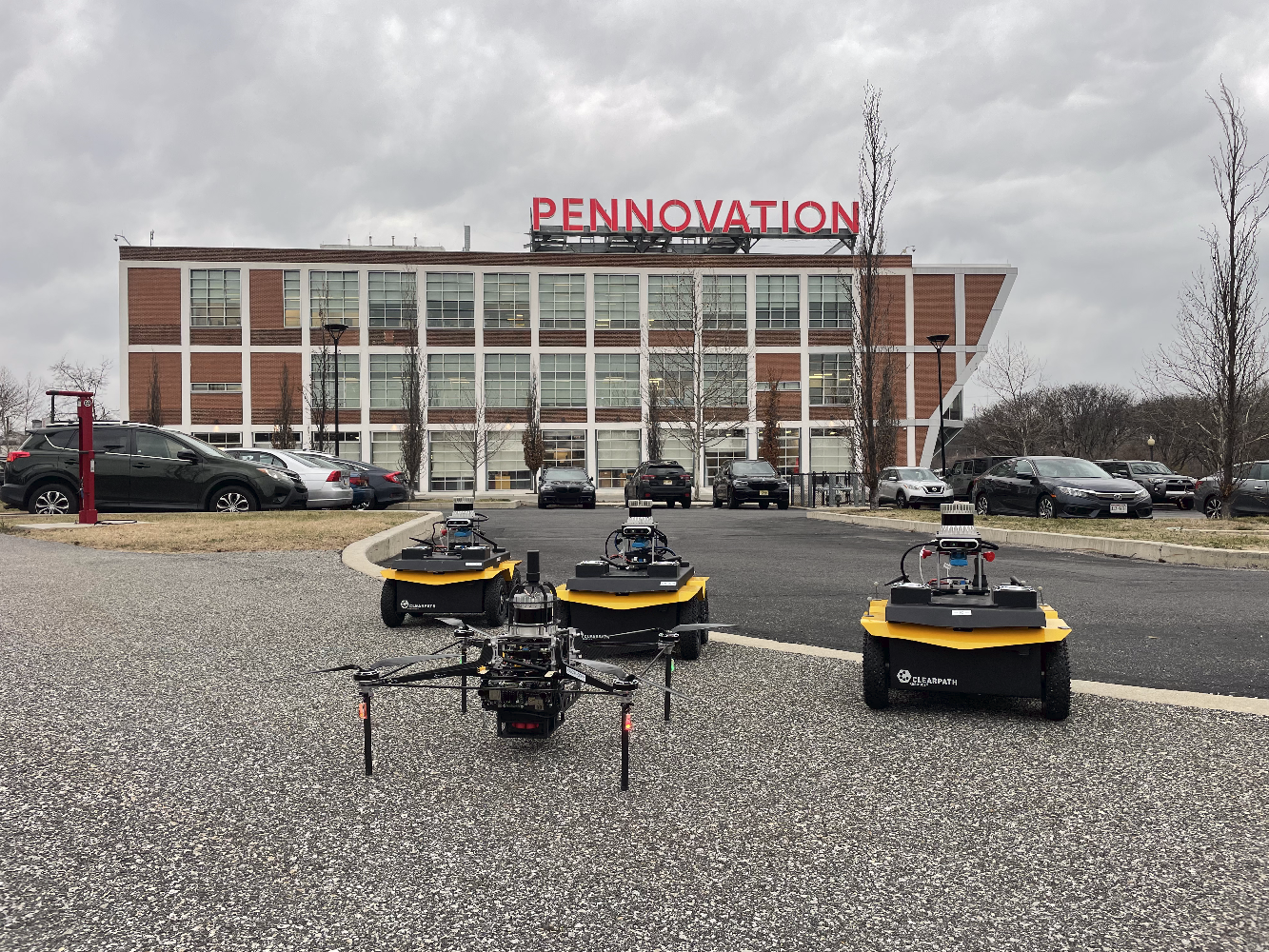}
     \caption{\textbf{UAVs and UGVs used in our experiments.}  
     Details and open-sourced autonomy stack of our UAVs can be found in \cite{liu2022large}, and details of our UGVs can be found in \cite{miller2022stronger}. Our UAVs carry lidars. However, our method can be used directly for lighter sensor packages such as stereo cameras and IMU.}
    \label{fig:robots}
    \vspace{-0.2in}
\end{figure}

We summarize our contributions as follows:
\begin{itemize}
    \item We propose a game theoretic formulation of the problem of positioning identical UGVs to enable robust global pose estimation of UAVs upon taking unlabelled measurements of UGVs.
    \item We present a hybrid analytic-search-based approach to solve the underlying nonconvex maximin optimization problem that uses exhaustive search in a low dimensional space in conjunction with analytic expressions to speed up computation.  
    \item We show that the number of ground vehicles necessary for global localization can be strictly larger when they are indistinguishable as opposed to when they have discernible perceptual identities.
    \item Through various experiments in the real world, we demonstrate that our algorithm (1) consistently outperforms random positioning of UGVs, (2) achieves centimeter-level global localization accuracy, and (3) can tolerate a significant amount of random or non-stochastic noise. 
\end{itemize}
 
A picture of the robot team used in our experiments is shown in \cref{fig:robots}. 
A demo video can be found at \url{https://www.youtube.com/watch?v=3_gXzmiNrRI}.

\section{Related Work}
\label{sec:related-work}

Examples of data association in robotics abound.
They range from recovering the 3D structure of the environment with a stereo camera to recognizing regions of space that can reduce the drift in odometry by e.g. performing loop closures. 
Data association can also involve relating observations captured by different robots. 
Overall, it is a challenging problem for several reasons. 
First, repetitive patterns together with sensor noise can diminish the efficacy of even the most discriminative feature descriptors, resulting in a large number of motion hypotheses consistent with sensory percepts.  
Second, data association typically relies on stationarity of some aspect of the environment (e.g. static objects), and unexpected behaviour may arise when the environment changes over time.
Some existing methods for data association develop algorithms for inferring optimal point estimates of the underlying geometry of the problem, whereas others allow for keeping track of multiple hypotheses consistent with sensory inputs.

State of the art approaches for recovering optimal point estimates exploit the invariance in the pairwise geometry of points inside a point cloud to perform robust data association and outlier rejection. A score for each pair of associations is computed based on the similarity of the resulting distances between the points. A consistency graph is built where each pair of associations is connected. The edge of the graph encodes association consistency. The largest set of consistent associations can then be identified by finding the maximum clique in this consistency graph. Some work uses techniques such as thresholding this graph into a binary graph \cite{mangelson2018pairwise-unweighted, yang2020teaser-unweighted, shi2021robin-unweighted}, which unavoidably leads to loss of information. Other works utilize a weighted graph \cite{leordeanu2005spectral-weighted, park2013fast-weighted}. However, this treatment leads to a violation of the clique constraint and ultimately leads to incorrect edges being added to the consistency graph and incorrect data association being chosen. CLIPPER \cite{lusk2021clipper} addresses these limitations by using edge weights while still maintaining clique constraints by selecting the densest clique of consistent associations and is shown to maintain higher precision than both unweighted and weighted benchmark methods. 
Other works also investigate the use of polygons as descriptors for place recognition and map merging \cite{nardari2020place}, where Urquhart tessellations are derived based on the positions of landmarks. Polygons are then computed and used to generate descriptors of the local neighborhood of the robot. The matching between current observations and target observations (e.g., a global map) is then carried out based on the polygon descriptors. 
However, point estimates are sometimes insufficiently expressive for accurate state estimation in perceptually aliased environments, especially when the level of sensor noise is high.

To reduce the impact of erroneous data association, state-of-the-art SLAM approaches incorporate data association by performing ``soft" inference of the state at each step \cite{bowman2017probabilistic, doherty2019multimodal} in the form of a weighted average of different hypotheses. 
However, the ambiguity in data association coupled with nonlinear measurement models typically leads to increasingly inaccurate relative hypothesis weights, which eventually leads to errors in state estimates.
Such errors can result in unsafe behaviour of highly agile aerial vehicles, and can degrade the quality of numerous downstream tasks, such as map merging.

We propose using a small number of carefully positioned robots as landmarks, and actively navigate them to form the least ambiguous ``constellations''.
The problem setting differs from that of our previous work \cite{spasojevic2023active}, which assumes that all robots can estimate their poses in the same reference frame and that data association is solved. 
In this work, the reference frame of the landmark robots is treated as the global frame. 
Their teammates do not have prior knowledge about the relative transformations between their reference frames and the global frame. 
They rely on the observation of the landmark robots to estimate both the data association and their poses in the global frame. 
The proposed algorithm also enables the aerial robot to find loop closures and perform map merging tasks in GPS-denied environments.

\section{Data Association Anticipative Positioning}

Let $\mathcal{L}$ be a set of ground robots with cardinality $M = \vert \mathcal{L} \vert$.
Suppose we require \textit{global} pose estimates of an aerial vehicle operating in a given region upon taking unlabelled collections of measurements of agents in $\mathcal{L}$. 
When estimating its pose from unlabelled measurements, the aerial agent will typically perform data association either before or jointly with its pose estimation. 
Intuitively, the former links every measurement to a physical agent (in $\mathcal{L}$) that induced it.  
In this way, we recover measurements to distinguishable landmarks, allowing the agent to readily triangulate its pose using available algorithms. 
However, mistakes in data association can result in gross errors in pose estimates. 
Our task involves positioning agents in $\mathcal{L}$ in order to maximize the robustness of data association at any point of the operating space of the aerial agent.
The aerial vehicle is informed of the positions of the agents in $\mathcal{L}$ before deployment.

\begin{figure}[t!]
\centering
\includegraphics[trim=100 0 80 0, clip,width=1.0\columnwidth]{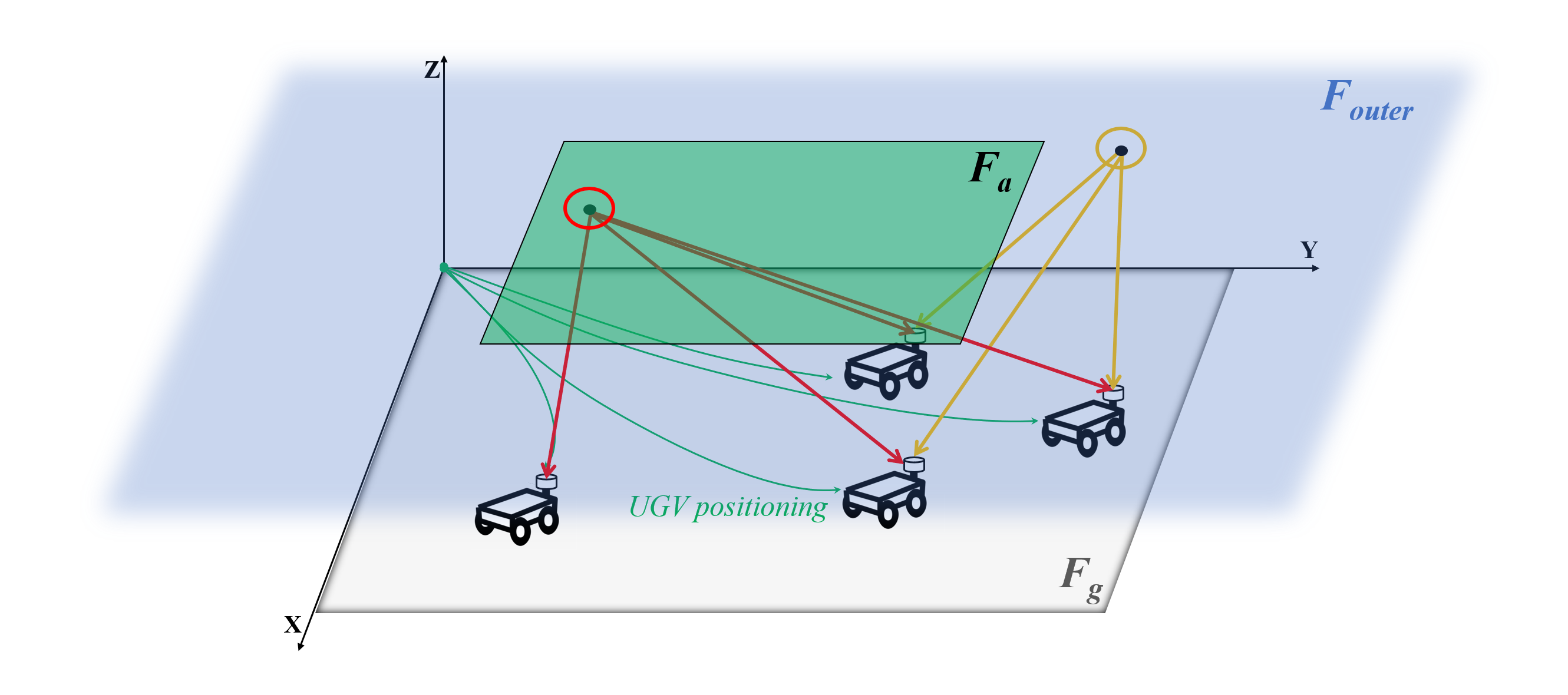}
    \caption{\textbf{Problem illustration.} Our goal is to position the UGVs in $\mathcal{F}_g$ (gray area) in such a way that we can distinguish any pair of distinct, suitably separated, robot poses in $\mathcal{F}_a$ (green region specified by the user) and $\mathcal{F}_{outer} \supset \mathcal{F}_a$ (blue region) using unlabelled measurements to the UGVs. Any UGV in $\mathcal{F}_g$ can be observed from any point in $\mathcal{F}_{a}$ - this assumption needs not hold for $\mathcal{F}_{outer}$. The red and orange lines illustrate two sets of measurements from two poses, red signifying observations from $\mathcal{F}_{a}$ while orange those from $\mathcal{F}_{outer}$ (not necessarily in $\mathcal{F}_{a}$). 
}
    \label{fig:problem-illustration-measurements}
    \vspace{-0.2in}
\end{figure}

In particular, suppose we have a function
\begin{equation} \label{eq:high_level_da_quality}
\begin{aligned}
\mathcal{Q} : (\mathbb{R}^3)^M & \rightarrow [0, \infty) \\
\mathcal{Q} : (z_{1:M}) & \mapsto [0, \infty)\\
\end{aligned}
\end{equation}
that measures the predicted quality of pose estimation with unlabeled measurements when the $\mathcal{L}$-agents are positioned at points $z_1, z_2, \dots, z_M$.
We solve 
\begin{problem} \label{prob:high_level_problem}
\begin{equation}
\begin{aligned}
\max_{z_{1:M}} \ & \mathcal{Q}(z_{1:M}) \\
& s.t. \ z_i \in \mathcal{F}_g \quad \forall \ i \in [M]. 
\end{aligned}
\end{equation}
\end{problem}
Here, $\mathcal{F}_g \subseteq \mathbb{R}^3$ denotes the region of space in which we can position ground agents. 
The function $\mathcal{Q}$, evaluating the merit of a positioning $z_{1:M}$, is defined as the value of the following
\begin{problem} \label{prob:merit_problem}
\begin{equation}
\begin{aligned}
\mathcal{Q}(z_{1:M}) & = \min_{(g_i)_{i = 1,2}, \Pi} \ || H(z_{1:M};\ g_1) - \Pi H(z_{1:M}; \ g_2) ||_2^2 \\ 
& \hspace{7mm} s.t. \quad \Pi \in S_{M} \\ 
& \hspace{15mm} g_1, \ g_2 \in SE(3) \\
& \hspace{15mm} || t_1 - t_2 ||_2 \geq R_{res} \text{ or }\\
& \hspace{20mm} \Big{\vert} \arccos{\frac{1}{2}(Tr(R_1^TR_2) - 1) \Big\vert \geq \beta_{res}} \\
& \hspace{15mm} t_1 \in \mathcal{F}_a, \ t_2 \in \mathcal{F}_{outer}.
\end{aligned}
\end{equation}
\end{problem}
Here, the function $H$ collects the \textit{sequence} of relative measurements from pose $g = (R, t)$ to points $z_1, z_2, \dots, z_M$:
\begin{equation}
H(z_{1:M}; \ g) = 
\begin{bmatrix}
h(g, z_1) \\
\vdots \\
h(g, z_M)
\end{bmatrix}.
\end{equation}
The pair $g_1 = (R_1, t_1)$ and $g_2 = (R_2, t_2)$ represents poses of hypothetical vantage points expressed with respect to the world frame - the first component representing orientation and the second  position. 
$\Pi$ is a (block-)permutation matrix, $\mathcal{F}_a$ is a region which the aerial agent should reach at least once, and $\mathcal{F}_{outer}$ encloses its operating space.
Last but not least, $R_{res}$ ($\beta_{res}$) roughly encodes the tolerated length scale of position (angular) uncertainty of the aerial agent when it is located inside $\mathcal{F}_a$.

In Problem \ref{prob:merit_problem}, we distinguish between regions $\mathcal{F}_{a}$ and $\mathcal{F}_{outer}$ for the following reason.
Requiring that the UAV remain in $\mathcal{F}_a$ at all times is an overly restrictive requirement.
We therefore allow it to take excursions away from the latter and into the much larger region $\mathcal{F}_{outer}$. 
During such excursions, the UAV forms an estimate of its global pose by stitching the odometry estimate accrued while moving through $\mathcal{F}_{outer}$ with its global pose estimate the last time it visited $\mathcal{F}_{a}$.
Before its first visit to $\mathcal{F}_{a}$, we assume the agent performs state estimation using odometry alone, and follows a trajectory which guarantees it reaches $\mathcal{F}_{a}$ eventually.

This naturally raises the question: how does an agent infer whether or not it is located in $\mathcal{F}_{a}$?
To answer this, we allow the UAV to attempt a ``new'' global pose estimation only once it registers measurements to $M$ UGVs (albeit in unknown order).
Tacitly, we assume that all UGVs are within sensing range at each point of $\mathcal{F}_a$.
Once a UAV receives a sufficient number of such measurements, it computes the data association and orientation yielding the smallest residual in the set of fitted measurements. 
If both the norm of the residual is small, and the mean position lies in $\mathcal{F}_a$, the agent keeps the latter as an estimate of its global pose - and discards it otherwise.

The rationale behind the definition of $\mathcal{Q}$ via Problem \ref{prob:merit_problem} is to allow the UAV agent (1) to tell whether or not it is located inside $\mathcal{F}_a$, and if so (2) where in $\mathcal{F}_a$ it lies. 
The permutation $\Pi$ and pose $g_2$ try to make measurements of ground agents at the latter vantage point as ``close as possible'' to corresponding measurements at $g_1$.
Our goal thus ends up being to find the positioning $z_{1:M}$ for which the possibility of ``confusing'' any such pair of poses, at a distance at least $R_{res}$ apart, is as small as possible. 
Physically, $\mathcal{Q}$ can be interpreted as the maximum amount of sensor noise which can be tolerated before such ambiguity arises. 
As a result, our problem is of a maximin nature. 
\section{Approach}
\label{sec:approach-method}

In this paper, the measurement function $h$ provides relative position (in the form of range + bearing) measurements to select objects in the environment. 
Namely,
\begin{equation} \label{eq:measurement_model}
h(g, z) \equiv h((R,t), z) = R^T(z - t) + \epsilon,
\end{equation}
where $\epsilon \in \mathbb{R}^3$ represents measurement noise. 
We also make three assumptions on pose estimates of the UAVs based on their sensor measurements excluding those modeled by $H$:
\begin{enumerate}
    \item they can be used to accurately estimate the pitch and roll, but not the yaw angle of the vehicle
    \item the altitude of the UAV is known at all times.
\end{enumerate}
In particular, the UAV is equipped with a high-quality IMU \cite{vectornav-dataset} that allows it to recover accurate estimates of its pitch and roll angles.
However, the remaining component of its orientation, the yaw angle, cannot be reliably obtained from the IMU due to magnetic interference induced by electric current drawn by the motors. 
Furthermore, lidar odometry alone cannot provide the absolute yaw angle of the vehicle w.r.t. the world frame.
The latter remains to be estimated. 
Last but not least, we assume the agent can get reliable estimates of its height, which can be easily acquired from onboard sensors such as a lidar or an altimeter.

Collecting the latter set of assumptions and performing suitable relative position transformations using known information, we may assume that all agents operate on a fixed plane parallel to the ground. 
Therefore, we solve a subfamily of Problems \ref{prob:high_level_problem} in which $\mathcal{F}_{a}$, $\mathcal{F}_{outer}$, and $\mathcal{F}_{g}$ are planar regions. 
We let $\mathcal{F}_{a}$ and $\mathcal{F}_{outer}$ be circles of radii $R_{a}$ and $R_{outer}$ centered on point $C_{a}$.
The region $\mathcal{F}_{g}$ is a rectangle centered at $C_{a}$ with half-diagonal $d_{semi}$ that satisfies $R_a + d_{semi} \leq R_{sense}$. 
$R_{sense}$ represents the sensing range of every UAV agent.
The former inequality ensures that no matter where we place a UGV in $\mathcal{F}_{g}$, it will be sensed from a vantage point of the UAV in $\mathcal{F}_{a}$.

\begin{algorithm}
\caption{Unlabelled UGV Positioning (U2GVP)} \label{alg:U2GVP}
\KwData{$\mathcal{F}_a$, $R_{res}$, $\mathcal{F}_{outer}$, $\mathcal{F}_{g}$}
\KwResult{$z_{1:M} \in \mathcal{F}_{g}$}
$z_{1:M}^{(0)} \gets$ initialize$(\mathcal{F}_{g})$\;
$z_{1:M} \gets$ ZerothOrder($z_{1:M}^{(0)}$, AdPC, $\mathcal{F}_a, R_{res}, \mathcal{F}_{outer}, \mathcal{F}_{g}$)\;
\end{algorithm}

\begin{algorithm}
\caption{Adversarial Pose Confounder (AdPC)} \label{alg:AdPC}
\KwData{$\mathcal{F}_a$, $R_{res}$, $\mathcal{F}_{outer}$, $z_{1:M} \in \mathcal{F}_{g}$}
\KwResult{$q \geq 0$}
$\Theta \leftarrow$ grid$((0,\ 2\pi))$\;
$q \gets M \times R_{res}^2$\;
$\bar{z} \gets \frac{1}{M} \sum_{i = 1}^{M} z_i$\;
$v_{aux} \gets 2 \sum_{i=1}^M ||z_i||_2^2 - 2M || \bar{z} ||_2^2$\;
\For{$\theta \in \Theta$}{
    $R \gets R(e_3;\ \theta)$\;
    $\rho_{rot} \gets M \times dist(R(\mathcal{F}_{a} - \bar{z}), \ \mathcal{F}_{outer} - \bar{z})^2$\;
    $\rho_{match} \gets 2 \times \max_{\sigma \in Sym(M)} \sum_{i=1}^M z_i^T R z_{\sigma(i)} $\;
    $q \gets \min\{ q, \ v_{aux} + 2M \bar{z}^TR\bar{z} - \rho_{match} + \rho_{rot} \}$\;
}
\end{algorithm}

Problem \ref{prob:high_level_problem} is a computationally challenging task. 
This is partially due to the fact that the function $\mathcal{Q}$ is specified as a solution to a non-convex minimization problem. 
For this reason, we adopt a hybrid exhaustive-local optimization procedure. 
Indeed, our high-level approach given in Algorithm \ref{alg:U2GVP} centers on refining a random initial guess of optimal positions of UGVs using a zeroth order optimization algorithm \cite{powell1994direct}, as implemented in \cite{nloptsuite}. 
The method in \cite{powell1994direct} requires us to supply values of the $\mathcal{Q}$ function at specified points, which we compute in Algorithm \ref{alg:AdPC} explained in more detail below.

Due to the non-convex nature of Problem \ref{prob:merit_problem}, we solve it using a combination of an exhaustive grid-based procedure and analytic computations which allow us to bypass further numerical optimization. 
Line 1 of Algorithm \ref{alg:AdPC} samples a set of non-zero angles encoding the relative orientation angle between $g_1^{*}$ and $g_2^{*}$ in an optimal solution $(g_1^{*}, g_2^{*})$ to Problem \ref{prob:merit_problem}. 
Line 2 represents the value of the $\mathcal{Q}$ function in case the relative orientation between the latter two poses is the identity transformation. 
The loop in line 5 then iterates through all the possible angles in $\Theta$.
At each fixed angle $\theta$, we analytically determine the pair of positions $(t_1, t_2)$ in $\mathcal{F}_a \times \mathcal{F}_{outer}$ at which the \textit{sets} of measurements of UGVs differ by as little as possible. 
The distance between a set of measurements is captured by the solution of a bipartite maximum weight matching problem in line 8. 
Even though the latter can be solved using polynomial time algorithms, for a small number of agents $M = \mathcal{O}(1)$, we can solve it by iterating through all permutations with low computational cost.
\section{Analysis}

\subsection{Analytic Solution}

In this section, we provide further details on our analytic solution of the problem in the loop of Algorithm \ref{alg:AdPC}.
We define the stacked vector of UGV positions via $Z = [z_1^T, z_2^T, \dots, z_M^T]^T$, and let $J = 1_M \otimes I_3$, where $1_M \in \mathbb{R}^M$ is a vector consisting of all ones.
The task inside the nested loop therefore involves solving the following problem:
\begin{equation} \label{eq:inner_loop_problem}
\min_{\substack{t_1 \in \mathcal{F}_{a}, \ t_2 \in \mathcal{F}_{outer}; \\ \Pi \in Sym(M)}} 
\left|\left| 
(\Pi \otimes I_3) (I_M \otimes R^T) (Z - J t_1) - (Z - J t_2)
\right|\right|_2^2.
\end{equation}
We start by noting that $\Pi$ is orthogonal since it is a permutation matrix. 
As a result, we have 
\begin{equation}
\begin{aligned}
(\Pi \otimes I_3) (\Pi \otimes I_3)^T & = (\Pi\Pi^T) \otimes I_3 = I_M \otimes I_3 = I  \\
(I_M \otimes R^T) (I_M \otimes R^T)^T & = I_M \otimes (R^T R) = I_M \otimes I_3 = I.
\end{aligned}
\end{equation}
The latter relation allows us to simplify the objective in Equation \ref{eq:inner_loop_problem}, via the following relations 
\begin{equation} \label{eq:simplified_inner_loop}
\begin{aligned}
& \left|\left| (\Pi \otimes I_3) (I_M \otimes R^T) (Z - J t_1) - (Z - J t_2) \right|\right|_2^2 = \\
|| Z - J t_1 ||_2^2 & + || Z - J t_2 ||_2^2 - 2 (Z - Jt_2)^T (\Pi \otimes R^T) (Z - J t_1) = \\
|| Z ||_2^2 - & 2M \bar{z}^T t_1  + M || t_1 ||_2^2 + || Z ||_2^2 - 2M \bar{z}^T t_2 + M || t_2 ||_2^2 \\
 - 2 (Z^T & (\Pi \otimes R^T) Z - M t_2^T R^T \bar{z} - M \bar{z}^T R^T t_1 + M t_2^T R^T t_1 ) =  \\
& 2 (|| Z ||_2^2 - Z^T (\Pi \otimes R^T) Z - M || \bar{z} ||_2^2 + M \bar{z}^T R^T \bar{z}) \\
 + M ||t_1 & - \bar{z} ||_2^2 + M ||t_2 - \bar{z} ||_2^2 - 2M (t_1 - \bar{z})R^T(t_2 - \bar{z}) = \\
 & 2 (|| Z ||_2^2 - Z^T (\Pi \otimes R^T) Z - M || \bar{z} ||_2^2 + M \bar{z}^T R^T \bar{z}) \\
 & + M || R(t_1 - \bar{z}) - (t_2 - \bar{z})||_2^2.
\end{aligned}
\end{equation}
To arrive at the second equality, we used 
\begin{equation}
    J^T Z = M\bar{z}, \ J^T \Pi = J^T, \ \Pi J = J,
\end{equation}
and finally in the third equality, we used the fact that $R$ is an orthogonal (in particular 2-norm preserving) matrix so that 
\begin{equation}
\begin{aligned}
M ||t_1 & - \bar{z} ||_2^2 + M ||t_2 - \bar{z} ||_2^2 - 2M (t_1 - \bar{z})R^T(t_2 - \bar{z}) = \\
M ||t_1 & - \bar{z} ||_2^2 + M ||R^T(t_2 - \bar{z}) ||_2^2 - 2M (t_1 - \bar{z})R^T(t_2 - \bar{z}) = \\
M || (t_1 & - \bar{z}) - R^T(t_2 - \bar{z}) ||_2^2 = M || R(t_1 - \bar{z}) - (t_2 - \bar{z}) ||_2^2.\\
\end{aligned}
\end{equation}
We can therefore read off two things from the last expression in Equation \ref{eq:simplified_inner_loop}.
Firstly when $\theta = 0$, corresponding to the case $R = I_3$, the expression becomes 
\begin{equation}
    2(|| Z ||_2^2 - Z^T (\Pi \otimes I_3) Z) + M || t_1 - t_2 ||_2^2,
\end{equation}
which, owing to the fact that $\Pi \otimes I_3$ is an orthogonal (2-norm preserving) matrix, is by the Cauchy-Schwartz inequality minimized by setting $\Pi \rightarrow I_M$, and choosing any $t_1 \in \mathcal{F}_{a}$, and $t_2 \in \mathcal{F}_{outer}$ so that $|| t_1 - t_2||_2 = R_{res}$. 
The latter observation justifies the assignment in line 2 of Algorithm \ref{alg:AdPC}. 
Second, when $\theta \neq 0$ and letting $R = R(\theta)$, the minimum of the objective in Equation \ref{eq:inner_loop_problem} can be obtained by separately minimizing 
\begin{equation}
|| R(t_1 - \bar{z}) - (t_2 - \bar{z}) ||_2^2
\end{equation}
over the set $(t_1, t_2) \in \mathcal{F}_{a} \times \mathcal{F}_{outer}$ and minimizing
\begin{equation}
-Z^T (\Pi \otimes R^T) Z = -\sum_{i=1}^M z_{\sigma(i)}^T R z_i
\end{equation}
over the set of all permutations $\sigma \in Sym(M)$. 
Clearly, the former is just the squared distance between sets $R(\mathcal{F}_{a} - \bar{z})$ and $\mathcal{F}_{outer} - \bar{z}$, whereas the latter amounts to solving a maximum weight matching problem, thus justifying lines 7 and 8 of the algorithm. 
Even though such matching can be found in polynomial time (in fact $\mathcal{O}(M^3)$), here we use an exhaustive procedure due to the budget-motivated assumption that $M = \mathcal{O}(1)$.
The running time of Algorithm \ref{alg:AdPC}, as we described, is then $\mathcal{O}(|\Theta|M!)$. 

\subsection{Impossibility Results}

\begin{figure}[t!]
\begin{center} 
\includegraphics[trim=0 20 0 20, clip,width=0.3\textwidth]{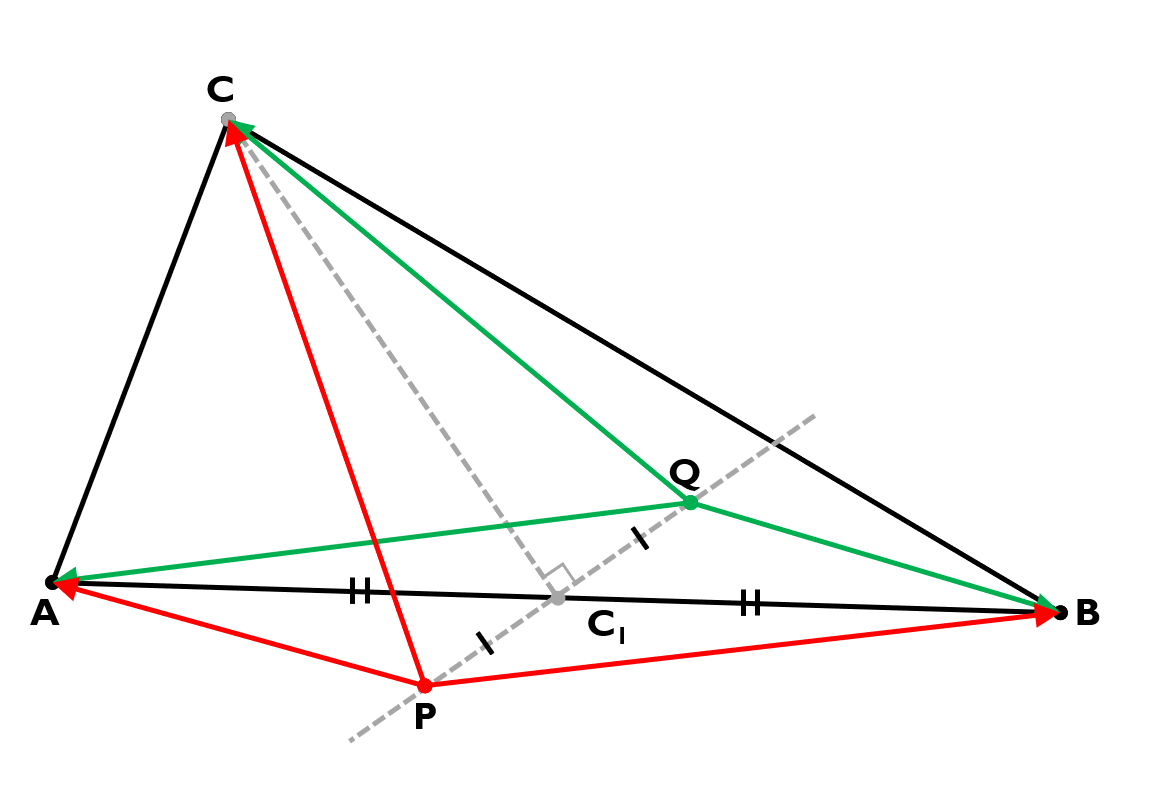}
\end{center}
\caption{\textbf{Illustration of perceptual aliasing with range-only measurements} to three unlabelled UGVs. The set of range measurements from points $P$ and $Q$ to points $A,B,$ and $C$ are the same. This follows from the fact that $APBQ$ is a parallelogram and that $\triangle CPQ$ is isosceles.}
\label{fig:range_only_impossibility}
\end{figure}

\begin{figure}[h!]
\begin{center} 
\includegraphics[trim=20 10 20 10, clip,width=0.4\textwidth]{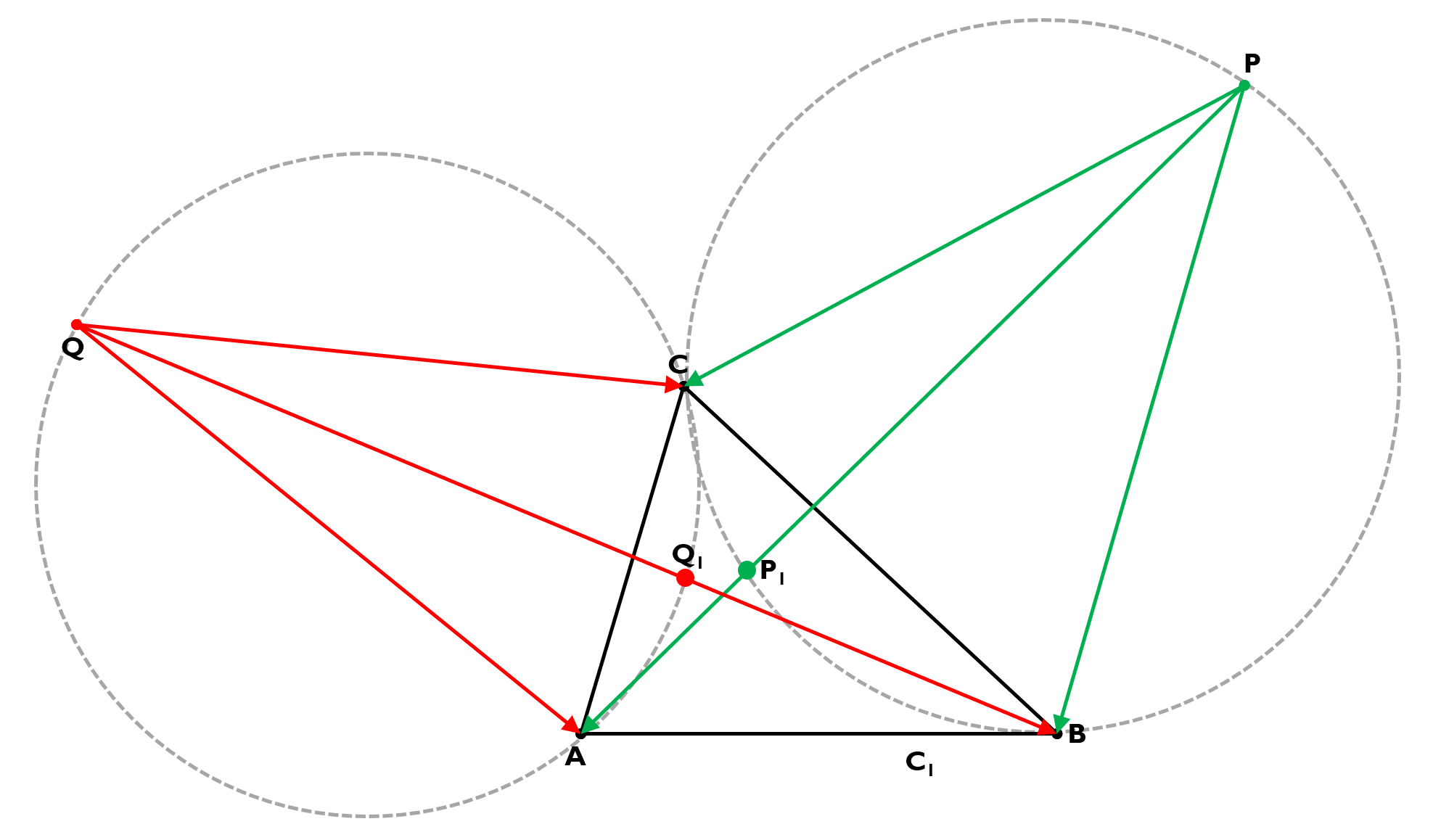}
\end{center}
\vspace{-0.1 in}
\caption{\textbf{Illustration of perceptual aliasing with bearing-only measurements} to three unlabelled UGVs. The set of oriented differential bearing measurements from points $P$ and $Q$ to points $A,B,$ and $C$ are the same. This follows from the fact that (1) $P$ and $Q$ have been chosen so that $\measuredangle CPB = \measuredangle AQC$; and (2) $P_1$ and $Q_1$ have been chosen so that $\wideparen{CP_1} / \wideparen{P_1B} = \wideparen{AQ_1} / \wideparen{Q_1C}$.
}
\label{fig:bearing_only_impossibility}
\end{figure}

In this section, we show that if we remove either bearing or range information from the set of unlabelled measurements, $M = 3$ ground agents no longer suffice. 
Note that if the UGVs were perceptually distinguishable, the latter would not be the case.
We believe these are results of independent interest, summarized as Propositions \ref{pop:range_only_impass} and \ref{pop:bearing_only_impass} with proofs sketched in Figures \ref{fig:range_only_impossibility} and \ref{fig:bearing_only_impossibility}, respectively.

\begin{proposition} \label{pop:range_only_impass}
Consider any three points $A, B, C$ in the plane. 
Then, there exists a pair of different positions $P, Q$ at which the sets of unlabelled range (i.e. distance) measurements to $A, B,$ and $C$ are the same. 
\end{proposition}

\begin{proposition} \label{pop:bearing_only_impass}
Consider any three points $A, B, C$ in the plane. 
Then, there exists a pair of different positions $P, Q$ at which the sets of unlabelled oriented (i.e. counterclockwise) bearing measurements to $A, B,$ and $C$ are the same. 
\end{proposition}

Finally, we give an intuitive explanation for why $M = 3$ UGVs for range and bearing suffice even with a priori unknown yaw information and data association. 
Indeed, consider placing the UGVs in an arbitrary scalene triangle. 
The unique lengths of its edges allow the agent to correctly estimate the identity of each of the three UGVs, and therefore recover its position and yaw with respect to the global frame.
\section{Experiments}
\label{sec:experiments}

\begin{figure}[t!]
\centering
\includegraphics[trim=100 150 150 230, angle=-0.5, clip,width=1.0\columnwidth]{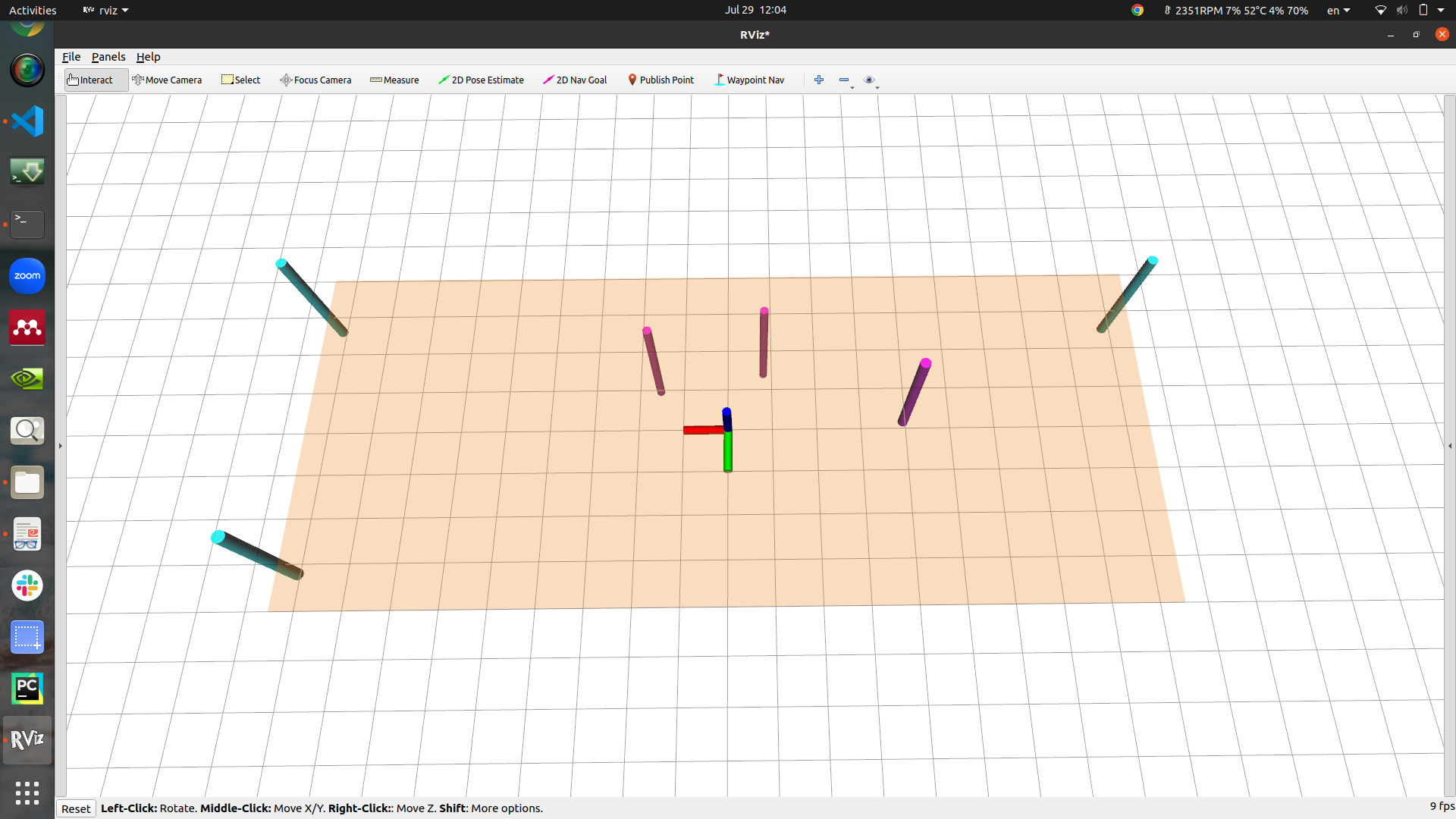}
\vspace{-0.2in}
    \caption{\textbf{Illustration of UGV positioning.} Each grid is 1 m $\times$ 1 m. 
    The cyan-colored cylinders visualize the optimal positioning of UGVs, and the magenta-colored cylinders visualize the random positioning of UGVs. The orange-colored region indicates the 8 m $\times$ 19 m  $\mathcal{F}_g$ region.}
    \label{fig:ugv-optimal}
\vspace{-0.2in}
\end{figure}

\begin{figure}[b!]
\begin{subfigure}{1.0\columnwidth}
\begin{center} 
\includegraphics[trim=0 128 0 0, clip,width=1.0\columnwidth]{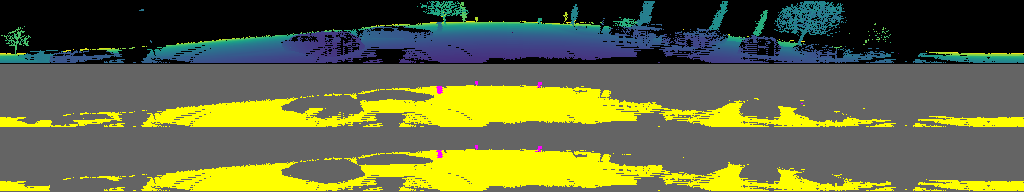}
\end{center}
\end{subfigure} 

\vspace{0.05 in}

\begin{subfigure}{1.0\columnwidth}
\begin{center} 
\includegraphics[trim=0 64 0 64, clip,width=1.0\columnwidth]{img/segmentation-example-better.png}
\end{center}
\end{subfigure} 

\vspace{0.05 in}

\begin{subfigure}{1.0\columnwidth}
\begin{center} 
\includegraphics[trim=0 0 0 128, clip,width=1.0\columnwidth]{img/segmentation-example-better.png}
\end{center}
\end{subfigure} 
    \caption{\textbf{Semantic segmentation example on validation set}. Top: lidar range image \xu{(color indicates intensity values). Middle: semantic segmentation (magenta: UGV, yellow: ground, grey: background). Bottom: ground truth. The UGVs and ground planes can be reliably segmented.}}
    \label{fig:ugv-seg}
\end{figure}

We evaluate our method's performance in real-world experiments with a UAV and a team of UGVs as illustrated in \cref{fig:robots}. 
We carry out three sets of experiments with different noise regimes: (1) real-world noise, (2) real-world noise with additional random noise, and (3) real-world noise with additional non-stochastic perturbation. 
The goal of (1) is to evaluate our system's performance and robustness with real-world noise, and the performance gain we can obtain over the random positioning of UGVs. 
The goal of (2) is to evaluate our method's robustness and performance margin over random positioning when the noise level increases. 
This can offer us insights into how our system can benefit robots with noisy sensors, such as depth cameras. 
The goal of (3) is to evaluate the robustness to the perturbation of the landmarks (i.e., if the UGVs are not positioned precisely at the places they should be, or if they get displaced by external forces). 
This allows us to evaluate how well our method performs under non-stochastic conditions. The optimal and random positioning of the UGVs is shown in \cref{fig:ugv-optimal}.

\begin{figure}[t!]
\begin{center} 
\includegraphics[trim=20 0 0 0, clip,width=0.5\textwidth]{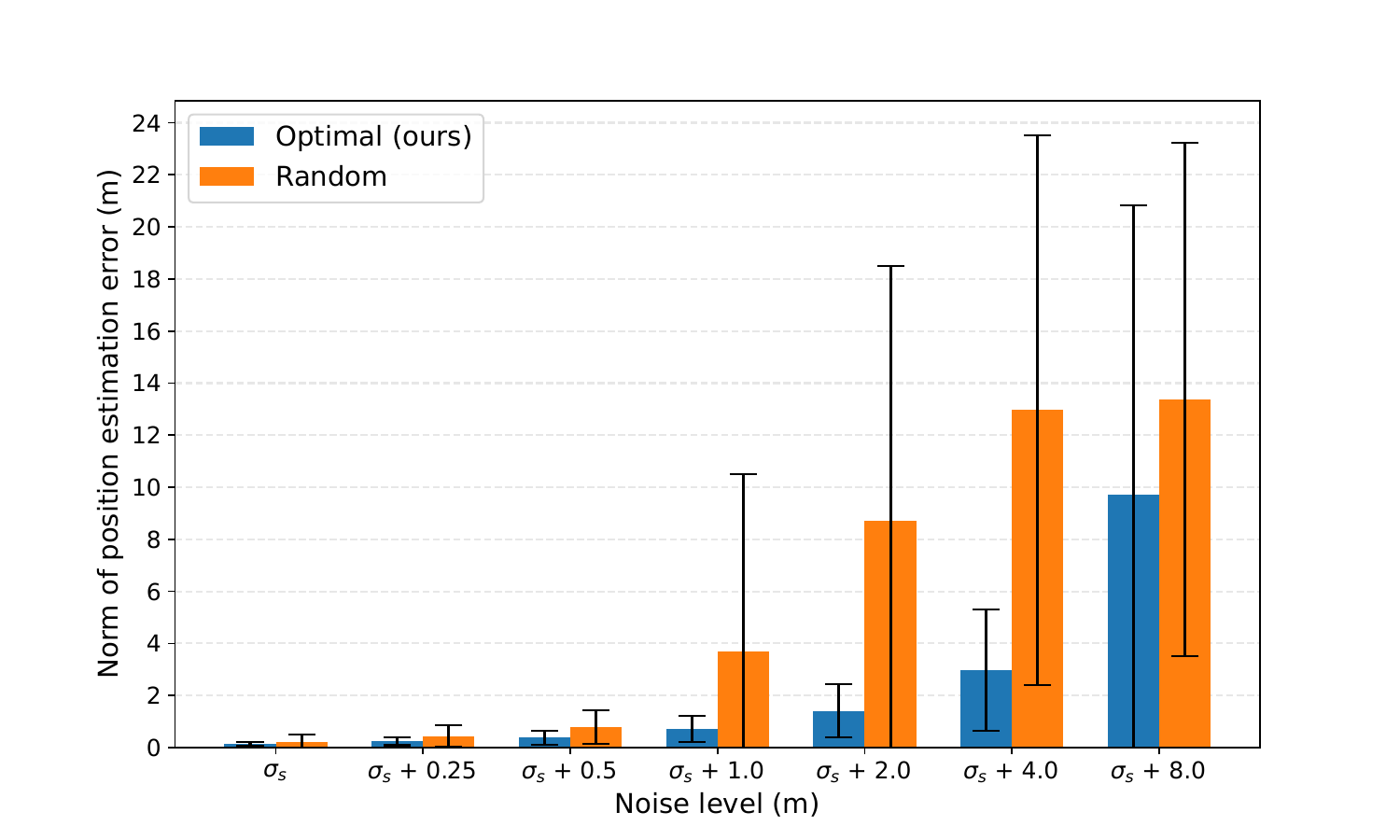}
\end{center}
\vspace{-0.1 in}
\caption{\textbf{Error in XY position estimates vs intensity of random noise.} The Y axis corresponds to the norm of the position error.   
The X axis represents the intensity of measurement noise $\sigma$ (in meters), where $\sigma_s$ denotes the level of nominal  noise.
We assume the UAV has reliable estimates of height above the ground since it has a lidar (this information is also readily available when using an altimeter). 
See \cref{sec:random vs non-stochastic} for details on how the noise is introduced and how ``random'' is different from ``non-stochastic''.  
With real-world noise, our method can position the UGVs in a configuration so as to localize the UAV accurately, with an error standard deviation of 7 cm and an error mean of 14 cm. 
Unlike odometry, our estimation errors will not accumulate with time since the UAV is localized w.r.t. a global reference frame instead of its previous pose.
Compared to random positioning, our method consistently reduces the error mean and standard deviation. Across all noise levels, the median reduction in error mean and error standard deviation is 51.63\% and 73.70\%, respectively.}
\vspace{-0.5em}
\label{fig:quant-result-table-error-xy-random}
\end{figure}

\subsection{Experiment setup}
\label{sec:exp-setup}
The experiment environment is shown in \cref{fig:robots}. Since the UGVs appear small in lidar scans, we add the same cylindrical payload to each UGV to augment detection. 
We use our modified lightweight version \cite{liu2022large} of RangeNet++ \cite{milioto2019irosrangenet++} to segment points into multiple classes in real time on board the UAV. 

From the segmented point clouds, we cluster the UGV points using DBSCAN \cite{ester1996density} to generate UGV detection instances. We then track such UGV detection instances across lidar scans. 
Once the UGV detection is robust (i.e., it has been tracked in multiple scans), we generate corresponding range and bearing measurements derived from this UGV detection. 
We estimate the ground plane by fitting a plane to all ground points.

Note that it is difficult to guarantee that each time step we can detect all UGVs. 
If the number of detected UGVs in a lidar scan is less than the total number of UGVs positioned in the environment, our method will skip it.
The $\mathcal{F}_a$ region is a circle with a diameter of 30m. The trajectory length of each UAV flight is around 200$\sim$300 meters. We position the UGVs to the corresponding locations in the global reference frame based on lidar odometry.

\subsection{Performance metrics}
\label{sec:performance-metrics}

The key metric for evaluating our system's performance is the UAV pose estimation accuracy. As explained in \cref{sec:approach-method}, we are concerned about estimating the UAV's XY position and yaw angle based on its observations of UGVs.

We use the state-of-the-art lidar-inertial odometry algorithm called faster-lio \cite{bai2022faster} to collect ground truth on the UAV poses. 
The drift of the faster-lio algorithm across all flights was under 10 cm. 
During the experiments, we treat the UAV's initial take-off pose as the origin of the global reference frame, and regard the pose estimated by lidar odometry as the ground truth when evaluating global localization accuracy. 
We sample key poses at 2-meter intervals. 
Across all samples we obtain (100$\sim$150 samples per flight), we calculate the error statistics between our estimated pose and the lidar odometry estimated pose. 
Finally, we calculate the mean and standard deviation of these errors and use them as performance metrics.

\begin{figure}[t!]
\begin{center} 
\includegraphics[trim=20 0 0 0, clip,width=0.5\textwidth]{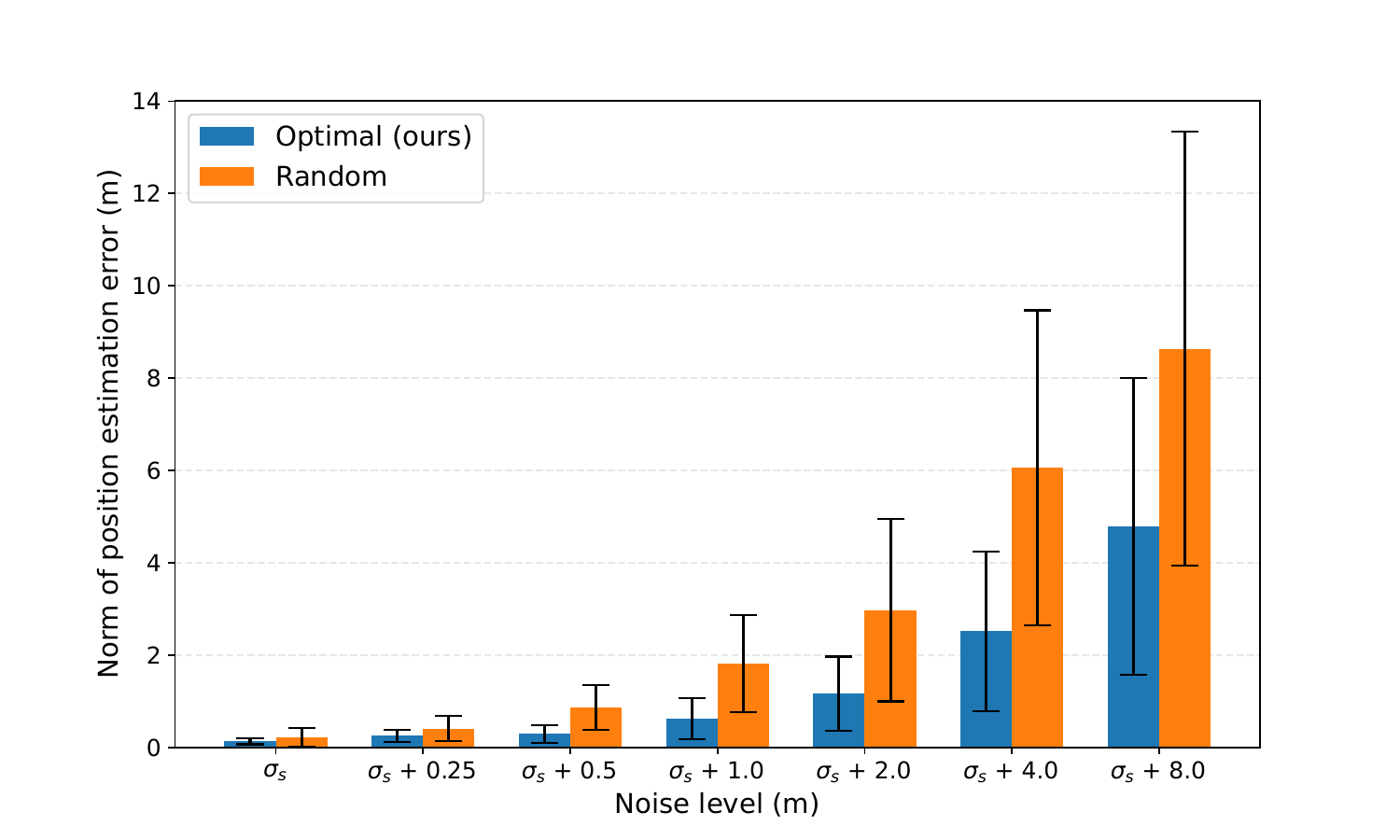}
\end{center}
\vspace{-0.1 in}
\caption{\textbf{Error in XY position estimates vs intensity of non-stochastic noise.} This figure follows the axis conventions of \cref{fig:quant-result-table-error-xy-random}. Compared to random positioning of UGVs, our method consistently and significantly reduces the error mean and standard deviation of position estimates of the UAV. Across all intensities of noise, the median reduction in error mean and error standard deviation is 58.46\% and 56.64\%. }
\vspace{-0.5em}
\label{fig:quant-result-table-error-xy}
\end{figure}

\begin{figure*}[t!]
\begin{subfigure}{0.49\textwidth}
\begin{center} 
\includegraphics[trim=20 0 0 0, clip,width=1.0\textwidth]{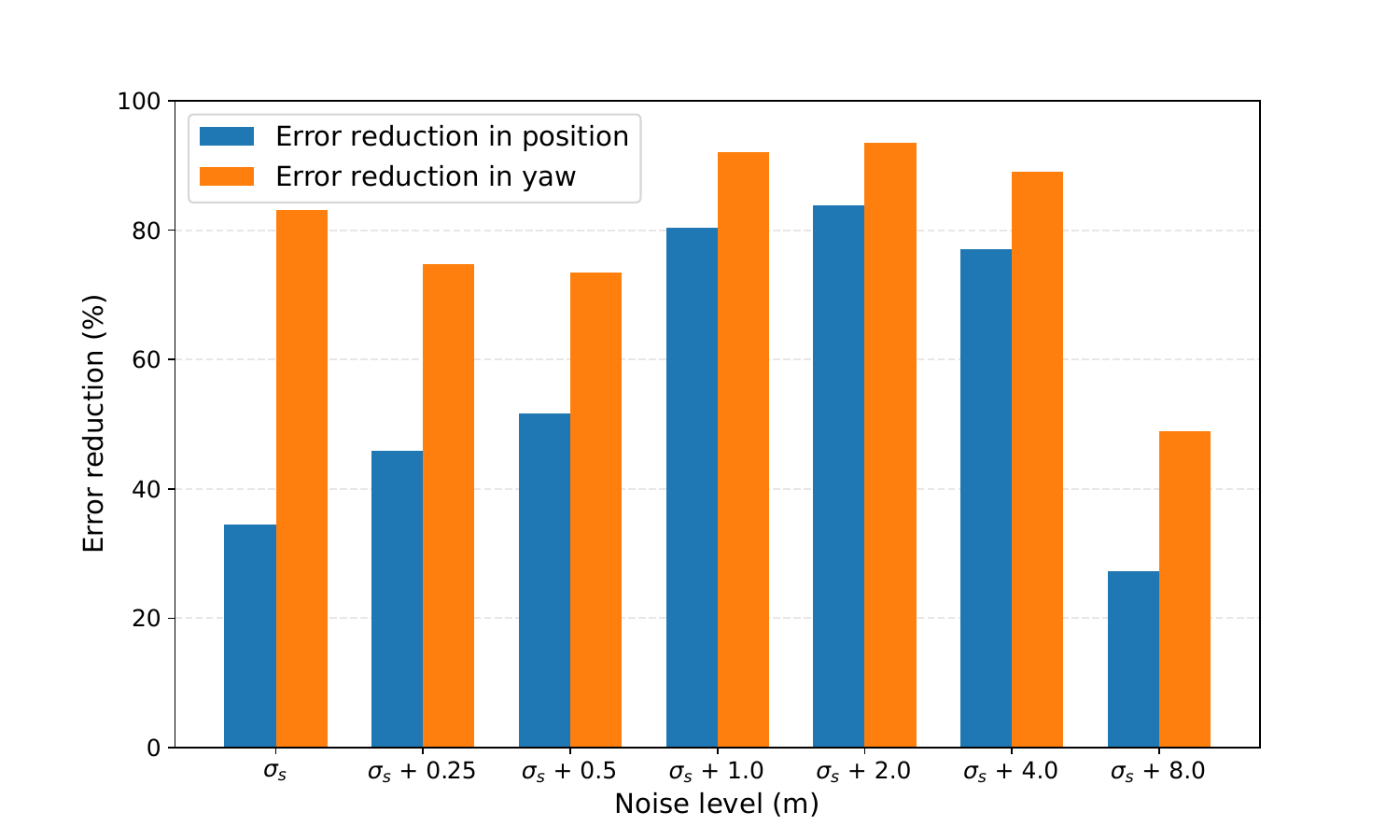}
\end{center}
\end{subfigure} 
\begin{subfigure}{0.49\textwidth}
\begin{center} 
\includegraphics[trim=20 0 0 0, clip,width=1.0\textwidth]{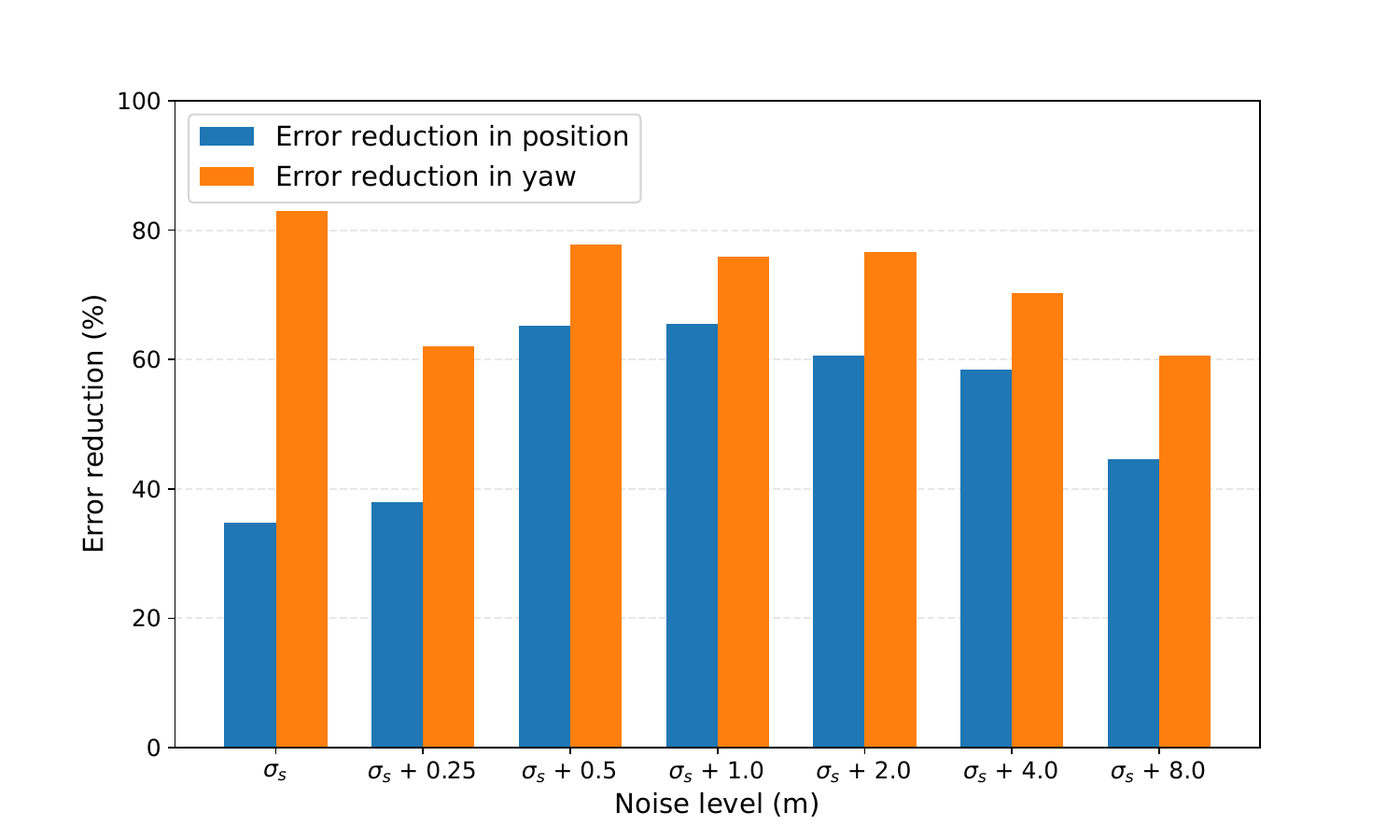}
\end{center}
\end{subfigure} 
\caption{\textbf{Percentage of reduction in error mean of XY position and yaw estimates v.s. intensity of random noise (left) and under non-stochastic noise (right).}  The Y axis is the percentage of reduction (i.e. $\frac{|\text{ours - random}|}{\text{ random}} \times 100$). The X axis is the noise level $\sigma$ (in meters). The margin of improvement in random noise is larger than in non-stochastic noise.} 
\label{fig:quant-result-table-error-reduction-random-and-non-stochastic}
\vspace{-0.5em}
\end{figure*}

\subsection{Random vs. non-stochastic noise}
\label{sec:random vs non-stochastic}
As mentioned above, we evaluate the performance of our system with different levels of noise. 
To achieve this, we first consider the effects of real-world noise (caused by noisy sensing and object detection). 
Then, in addition to such real-world noise, we add different levels of Gaussian noise to the position of each UGV. 
In other words, all studies have noise levels \textbf{no less than} than real-world noise. 
Two sets of studies are carried out: Random and Non-stochastic.

In both sets of studies, we first detect and localize the UGVs in the robot body frame. 
Then we use pose estimates from lidar odometry to transform these UGV observations into the world frame, where we add noise to UGVs' observed positions.
Finally, we generate the range-bearing measurements based on the relative position of the noisy UGV observations and the UAV.

In the random case, we add a different Gaussian noise to each UGV's position at each time step. Therefore, the noise added to the positions of the UGVs in the global frame varies with time. In the non-stochastic case, we first sample M different Gaussian noise values (M = number of UGVs), and use this set of noise values for all time steps. 
Effectively, this adds a fixed perturbation to each UGV's position in the global frame. 
This can happen when the UAVs fail to reach the desired positions precisely, or they are displaced by external forces.

\subsection{Quantitative results and analysis} 
\label{sec:quant-results}
\begin{figure}[t!]
\begin{center} 
\includegraphics[trim=20 0 0 0, clip,width=0.5\textwidth]{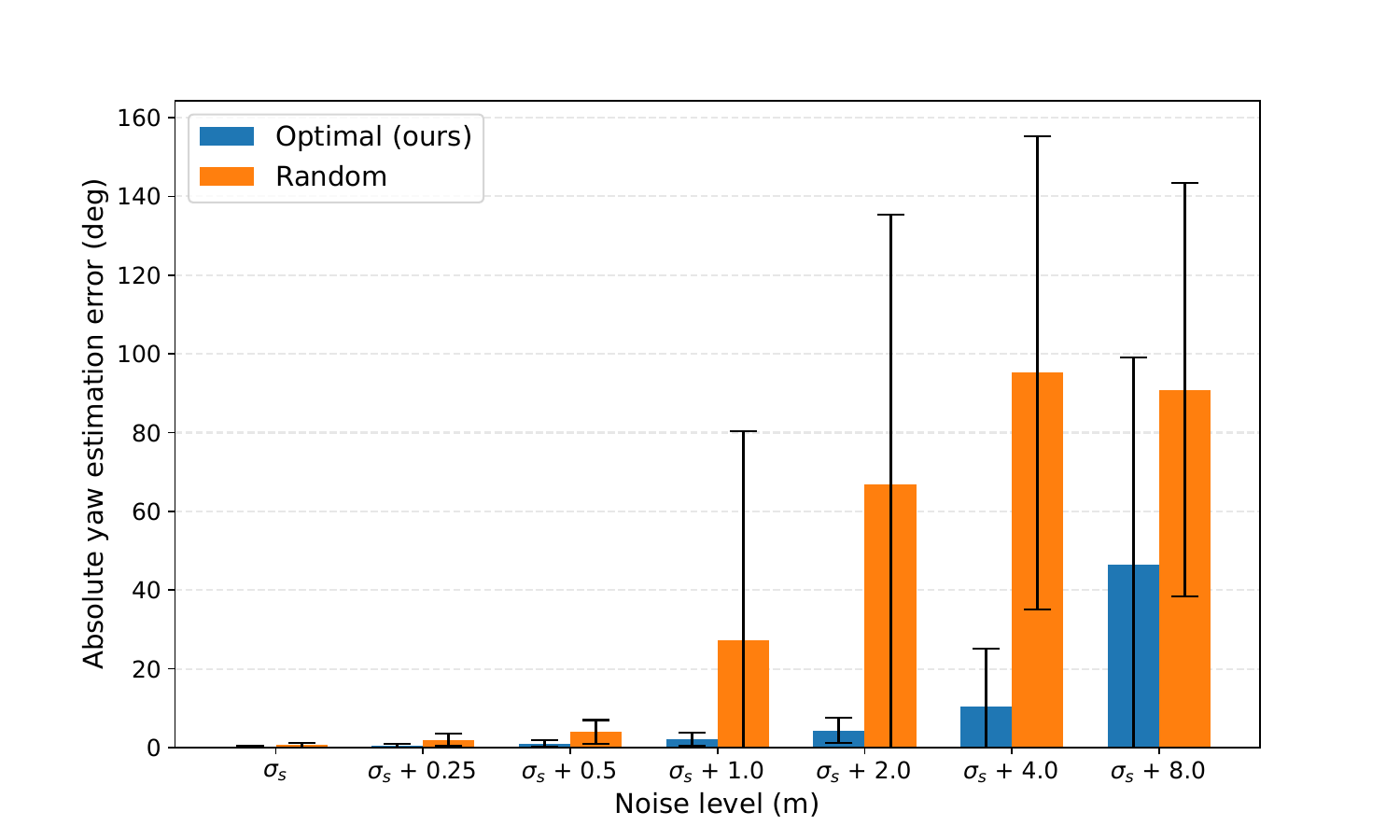}
\end{center}
\vspace{-0.5em}
\caption{\textbf{Error in yaw estimates vs intensity of random noise.} The Y axis represents the yaw error (in degrees). We assume the UAV has reliable estimates of roll and pitch angles since it runs state-of-the-art lidar odometry and has an IMU. 
The X axis is the noise level $\sigma$ (in meters). 
With real-world noise ($\sigma_{s}$), our method can position the UGVs in a configuration so as to estimate the UAV's heading accurately, with an error standard deviation of 0.27$^{\circ}$ and an error mean of $-0.09^{\circ}$. This is an order of magnitude more accurate than high-quality magnetometers.  For example, VectorNav's VN-100 \cite{vectornav-dataset} IMU's magnetometer can estimate heading up to  $2^{\circ}$ accuracy even in the ideal magnetic environments \cite{vectornav-heading-estimation},  which is usually not true for UAVs since they are susceptible to strong magnetic interference induced by the electric current drawn by motors. Meanwhile, the lidar odometry alone cannot estimate the absolute yaw angle w.r.t. the world frame. Across all noise levels, the median reduction in error mean and error standard deviation is 83.13\% and 71.65\%.}
\vspace{-0.5em}
\label{fig:quant-result-table-error-yaw-random}
\end{figure}

\begin{figure}[t!]
\begin{center} 
\includegraphics[trim=20 0 0 0, clip,width=0.5\textwidth]{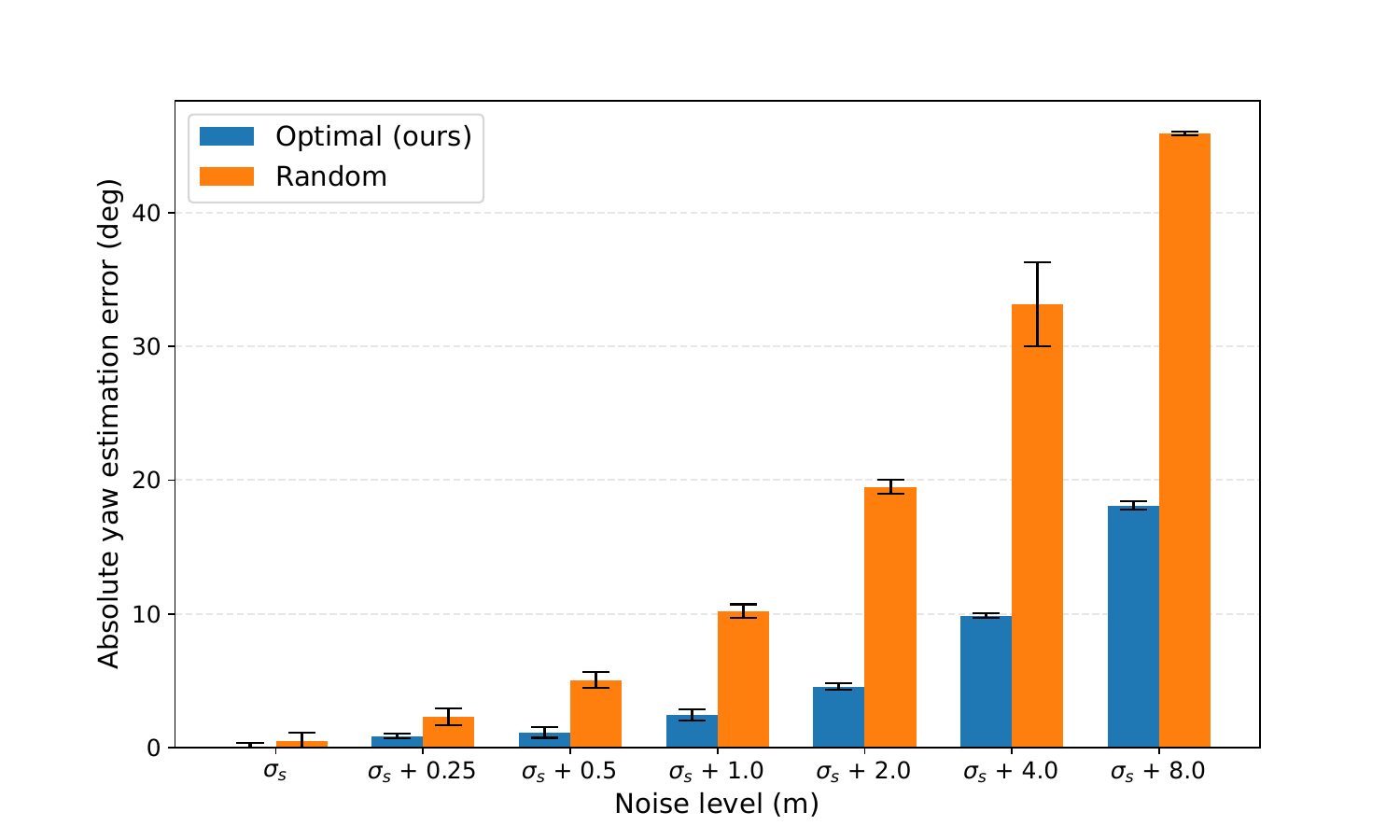}
\end{center}
\vspace{-0.5em}
\caption{\textbf{Error in yaw estimates under non-stochastic noise.} This figure follows the axis conventions of \cref{fig:quant-result-table-error-yaw-random}. Across all noise levels, the median reduction in error mean and error standard deviation is 75.95\% and 52.22\%.} 
\vspace{-0.5em}
\label{fig:quant-result-table-error-yaw-Adversarial}
\end{figure}

The results for X-Y position estimation are shown in \cref{fig:quant-result-table-error-xy-random} and \cref{fig:quant-result-table-error-xy}. Those for yaw angle estimation are shown in \cref{fig:quant-result-table-error-yaw-random} and \ref{fig:quant-result-table-error-yaw-Adversarial}. \cref{fig:quant-result-table-error-reduction-random-and-non-stochastic} illustrates the reduction of errors by our method. We summarize three significant findings from these results.

First, with real-world noise, our method outputs robust and accurate position and orientation estimates. The errors have a mean of 7 cm for the X-Y location and 0.28 degrees for the yaw angle, and a standard deviation of 14 cm for the X-Y location and 0.1 degrees for the yaw angle. This shows that by using a very small number (three) of unlabeled range and bearing measurements, we can reliably associate each measurement with each UGV at various viewing angles and ranges within the $\mathcal{F}_a$ region, and achieve high accuracy.

Second, with real-world as well as increased levels of noise, positioning of UGVs according to our algorithm consistently leads to more accurate state estimates than those derived from random positioning of UGVs. \cref{fig:quant-result-table-error-xy-random} shows that across a range of different levels of random noise, the median reduction of error mean and standard deviation of XY position estimates is 51.63\% and 73.70\%, respectively. 
Similarly, \cref{fig:quant-result-table-error-yaw-random} shows that the median reduction of error mean and standard deviation of yaw angle estimation is 83.13\% and 71.65\%, respectively.  
The corresponding results for non-stochastic noise are illustrated in \cref{fig:quant-result-table-error-xy} and \cref{fig:quant-result-table-error-yaw-Adversarial}.

Last but not least, the performance margin of our method over the random positioning increases with the level of noise. This is illustrated in \cref{fig:quant-result-table-error-reduction-random-and-non-stochastic}. In practice, the most common range-bearing sensors, such as lidars, usually have relative position measurements whose noise standard deviation is well below 2 m. For robot platforms that cannot afford to carry accurate, heavy, and expensive sensors such as lidars but can carry alternative range-bearing sensors such as stereo cameras, our method will bring even more significant improvement to the localization than just relying on randomly placed features (or natural landmarks). The same is true for environments with non-stochastic noise.

In summary, results from our real-world experiments show that our method can accurately estimate the pose. Furthermore, it is robust to significant levels of noise. 
Through comparisons against random positioning of UGVs, both the robustness and accuracy of pose estimation are shown to be drastically improved by positioning the UGVs using our method.

\section{Conclusion}
\label{sec:conclusion}

We address the problem of localizing UAVs in an environment without GPS and with very few natural landmarks. We illustrate the benefit of actively controlling identical ground robots that can serve as semantic landmarks that allow a UAV to align its local reference frame to the world frame.  
Our game theoretic formulation led us to a hybrid analytic-exhaustive-search algorithm for positioning the UGVs in a way that makes the problem of data association robust to a high level of noise. 
Our real-world experiments showed that our method achieves centimeter-level global localization accuracy regardless of the trajectory length, while only relying on observations of three actively controlled UGVs. 
Furthermore, our method was able to accommodate random as well as non-stochastic perturbations on measurements, indicating its potential for robust state estimation of SWaP-constrained UAVs.
This work enables robust multi-robot collaborative localization and mapping for multiple aerial robots in perceptually-challenged GPS-denied environments and without prior knowledge of relative transformation between the robots.
Future work will involve extending the algorithm to different measurement models, such as range-only or bearing-only sensors, and allowing UGVs to complement existing natural object landmarks in the environment to form a non-ambiguous constellation.

\bibliographystyle{IEEEtran}
\bibliography{ref}

\end{document}